\documentclass[10pt,twocolumn,letterpaper]{article}
\usepackage[pagenumbers]{cvpr}
\usepackage{graphicx}
\usepackage{amsmath}
\usepackage{amssymb}
\usepackage{booktabs}
\usepackage{multicol}
\usepackage{multirow}
\usepackage[pagebackref,colorlinks]{hyperref}
\usepackage[capitalize]{cleveref}
\crefname{section}{Sec.}{Secs.}
\Crefname{section}{Section}{Sections}
\Crefname{table}{Table}{Tables}
\crefname{table}{Tab.}{Tabs.}
\newcommand\blfootnote[1]{%
  \begingroup
  \renewcommand\thefootnote{}\footnote{#1}%
  \addtocounter{footnote}{-1}%
  \endgroup
}

\begin{document}
\title{Boosting Semi-Supervised 3D Object Detection with Semi-Sampling}
\author{Xiaopei Wu\textsuperscript{\rm 1,2*}, Yang Zhao\textsuperscript{\rm 1*}, Liang Peng\textsuperscript{\rm 1}, Hua Chen\textsuperscript{\rm 3}, Xiaoshui Huang\textsuperscript{\rm 2},\\ Binbin Lin\textsuperscript{\rm 1}, Haifeng Liu\textsuperscript{\rm 1}, Deng Cai\textsuperscript{\rm 1}, Wanli Ouyang\textsuperscript{\rm 2} \\
	\textsuperscript{\rm 1} Zhejiang University \quad
        \textsuperscript{\rm 2} Shanghai AI Laboratory \quad
	\textsuperscript{\rm 3}	Beijing Aircraft Technology Research Institute \\
	{\tt\small \{wuxiaopei, zhaoyang6, pengliang\}@zju.edu.cn}}
 
\maketitle

\begin{abstract}
    Current 3D object detection methods heavily rely on an enormous amount of annotations. Semi-supervised learning can be used to alleviate this issue. Previous semi-supervised 3D object detection methods directly follow the practice of fully-supervised methods to augment labeled and unlabeled data, which is sub-optimal. In this paper, we design a data augmentation method for semi-supervised learning, which we call \textbf{Semi-Sampling}.
    Specifically, we use ground truth labels and pseudo labels to crop gt samples and pseudo samples on labeled frames and unlabeled frames, respectively. Then we can generate a gt sample database and a pseudo sample database. When training a teacher-student semi-supervised framework, we randomly select gt samples and pseudo samples to both labeled frames and unlabeled frames, making a strong data augmentation for them.
    Our semi-sampling can be regarded as an extension of gt-sampling to semi-supervised learning. Our method is simple but effective. We consistently improve state-of-the-art methods on ScanNet, SUN-RGBD, and KITTI benchmarks by large margins. For example, when training using only 10\% labeled data on ScanNet, we achieve 3.1 mAP and 6.4 mAP improvement upon 3DIoUMatch in terms of mAP@0.25 and mAP@0.5. When training using only 1\% labeled data on KITTI, we boost 3DIoUMatch by 3.5 mAP, 6.7 mAP and 14.1 mAP on car, pedestrian and cyclist classes. Codes will be made publicly available at \url{https://github.com/LittlePey/Semi-Sampling}.
\end{abstract}

\blfootnote{*: equal contribution}

\vspace{-5mm}
\section{Introduction}
In recent years, the boom in deep learning has led to the rapid development of 3D object detection. Many 3D object detection methods \cite{liu2021group, misra2021end, rukhovich2021fcaf3d, shi2020pv, zheng2021se, wu2022sparse, bai2022transfusion, li2022deepfusion} make impressive performances. However, the 3D object detection task heavily relies on the availability of large 3D datasets annotated very well. The careful annotations for the raw data can be costly on both human resources and time. By contrast, the cost of acquiring unlabeled data is far less than labeled data,  and the amount of unlabeled data is huge. The natural idea is to exploit unlabeled data to facilitate model learning. Semi-supervised learning (SSL) provides means of leveraging unlabeled data to improve model performance when labeled data is limited. Existing semi-supervised 3D object detection methods usually use a teach-student mutual learning framework to facilitate student learning from the EMA teacher. SESS\cite{zhao2020sess} is the pioneering work. It designs a thorough perturbation scheme to enhance the generalization of the network on unlabeled data. 3DIoUMatch\cite{wang20213dioumatch} makes use of unlabeled data in the form of pseudo labels. It introduces a confidence-based filtering mechanism to remove poorly localized pseudo labels and achieves notable improvement compared to SESS.

In this work, we concentrate on data augmentation for semi-supervised learning. For the student network, the training data consists of two parts: labeled data and unlabeled data. Data augmentation for labeled data has been intensively studied, while data augmentation for unlabeled data is under-explored. 
For example, the gt-sampling~\cite{yan2018second} strategy is a practical data augmentation for labeled data. It crops ground truth samples from other frames and pastes them to empty space of the current frame with the help of collision detection. However, there is no its counterpart on unlabeled data. Considering that unlabeled data account for most of the training data in semi-supervised learning, it is meaningful to perform an object sampling strategy similar to gt-sampling on them. In addition, the massive unlabeled data can provide a vast amount of unseen objects, which can be cropped and pasted to both labeled or unlabeled frames, making a stronger data augmentation. 

Based on the above analysis, we propose a general object sampling strategy called \textbf{Semi-Sampling}.
The key idea is to bridge the object sampling gap between labeled frames and unlabeled frames.
We can use gt-sampling not only on labeled frames but also on unlabeled frames. We can crop objects not only in labeled frames but also in unlabeled frames for object sampling, which we call \textit{pseudo-sampling}. Our semi-sampling can be regarded as a general extension of gt-sampling in semi-supervised learning, and our method can increase the diversity of semi-supervised training data greatly. Extensive experiments on indoor and outdoor datasets demonstrate its effectiveness. 

To further boost the fully-supervised methods with semi-supervised learning, we construct a new dataset Omni-KITTI with KITTI raw data. KITTI raw data consists of many sequences, and the KITTI 3D object detection dataset is sampled from these sequences. Many frames in KITTI raw are not annotated, which can be used as unlabeled data. We provide a benchmark for Omni-KITTI, which gives the results of 3DIoUMatch and our method on Omni-KITTI.

To summarize, our contributions are listed as follows:
\begin{itemize}
\item We propose a new object sampling strategy \textbf{Semi-Sampling} to augment both labeled frames and unlabeled frames.
\item We achieve remarkable improvement over prior art on the indoor 3D object detection benchmark ScanNet, SUN-RGBD, and outdoor 3D object detection benchmark KITTI.
\item We contribute a new benchmark dataset setting called Omni-KITTI for semi-supervised learning, and we also achieve a promising result on the dataset.
\end{itemize}

\section{Related Works}

\subsection{Semi-Supervised Learning}
\textbf{Semi-Supervised Learning(SSL)} draws growing attention in a wide range of research areas like image classification and segmentation. Since unlabeled data can be obtained more easily than labeled data, unlabeled data is far more than labeled data. Semi-supervised learning focuses on leveraging the labeled and unlabeled data to help the learning of the model. Current semi-supervised methods can be roughly divided into two groups. One is pseudo-label based methods like \cite{lee2013pseudo}. The method pre-trains a model on labeled data and uses the pre-trained model to infer the unlabeled data. The prediction will be used as the ground truth of the unlabeled data. For the pseudo-label based method, setting a classification confidence threshold to filter pseudo labels with low quality can bring a big improvement to the performance, which is shown in FixMatch\cite{sohn2020fixmatch}. Another kind of method is based on consistency regularization, such as \cite{bachman2014learning,miyato2018virtual, sajjadi2016regularization,laine2016temporal}. They build a regularization loss with unlabeled images, which encourages the model to generate similar predictions on different perturbations of the same image. There are many ways to perform perturbations, such as conducting data augmentation on input data\cite{sajjadi2016regularization} and adding perturbations to the model\cite{bachman2014learning}. The methods mentioned above commonly use a framework called teacher-student, which is proposed in\cite{tarvainen2017mean}. In the teacher-student framework, the teacher network will be initialized with frozen weight, and the weight of the teacher will be upgraded during training as the EMA of the student netowrk.

\subsection{Semi-Supervised Object Detection}
Object detection is an essential task in computer vision, and great progress has been made in both 2D and 3D object detection areas. For semi-supervised object detection, they can also be separated into two groups: consistency methods \cite{jeong2019consistency, tang2021proposal} and pseudo-label methods \cite{sohn2020simple, liu2021unbiased, xu2021end, zhang2022semi}. CSD\cite{jeong2019consistency} proposes a consistency regularization to keep the consistency between predictions of an image and its flipped version. STAC\cite{sohn2020simple} adopts a two-stage scheme for training Faster R-CNN\cite{ren2015faster}. In the first stage, STAC pre-trains a detector with labeled data and takes the predictions of the pre-trained detector on the unlabeled data as pseudo labels. In the second stage, it selects pseudo labels with high confidence and updates the model by enforcing consistency with strong augmentations. 

For the semi-supervised 3D object detection in indoor scenes, SESS\cite{zhao2020sess} is based on the consistency loss method, and 3DIoUMatch\cite{wang20213dioumatch} is based on the pseudo-label method. SESS utilizes a mutual learning framework composed of a student and an EMA teacher. It uses three kinds of consistency losses to align the predictions of the student and teacher. 3DIoUMatch also uses a teacher-student framework, while it uses the predictions from the EMA teacher network as pseudo labels to supervise the student network on unlabeled data. For outdoor semi-supervised 3D object detection, most methods \cite{qi2021offboard, caine2021pseudo, park2022detmatch, xu2021semi} use the pseudo-label method. 3D Auto Labeling\cite{qi2021offboard} leverages a multi-frame detector and an object-centric auto-labeling model to generate accurate pseudo labels. 

\begin{figure*}[t]
	\vspace{-4pt}
	\begin{center}
		\setlength{\fboxrule}{0pt}
		\fbox{\includegraphics[width=0.85\textwidth]{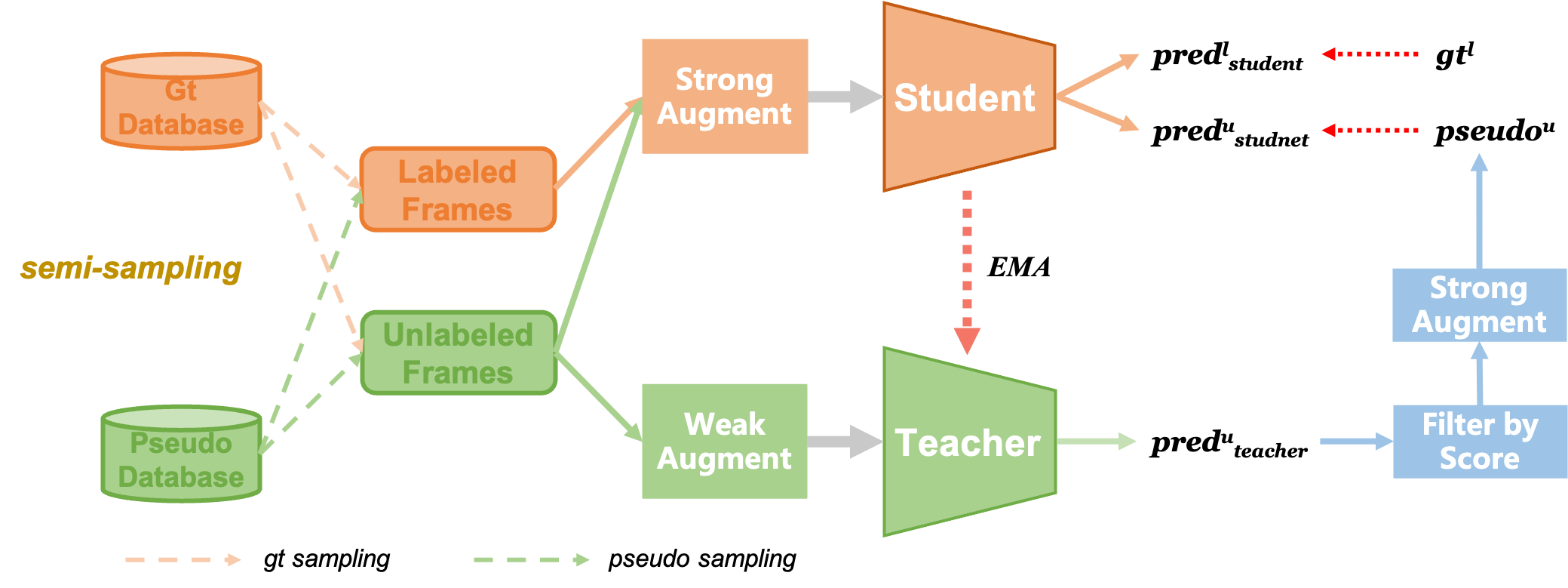}}
	\end{center}
	\captionsetup{font={small}}
	\vspace{-10pt}
	\caption{Illustration of our semi-sampling. We use ground truth bounding boxes to crop gt samples from labeled frames, generating a gt database. And we use pseudo labels on unlabeled frames to crop pseudo samples from unlabeled frames, generating a pseudo database. Semi-sampling augments the labeled and unlabeled frames by performing gt-sampling and pseudo-sampling on both of them.}
	\label{fig:overview}
	\vspace{-6pt}
\end{figure*}

\section{Methods}
\subsection{Problem Definition}
Given a scene $\boldsymbol{x}$ containing a set of objects that are represented as a 3D point cloud, where $\boldsymbol{x} \in \mathbb{R}^{n\times 3}$ denotes the scene containing $n$ points, our task is to localize the amodal oriented 3D bounding boxes for these objects and get their semantic class labels. Under the semi-supervised setting, the problem is more challenging due to the limited supervising information. Assuming that we have access to $N_l$ labeled point clouds and $N_u$ unlabeled point clouds. We donate the $i^{th}$ labeled point cloud as $P^l_i = (\boldsymbol{x}^l_i, \boldsymbol{y}^l_i)$ and the $i^{th}$ unlabeled point cloud as $P^u_i = \boldsymbol{x}^u_i$, where $\boldsymbol{x}^l_i$ and $\boldsymbol{x}^u_i$ are point clouds and $\boldsymbol{y}^l_i$ is ground truth annotations for $\boldsymbol{x}^l_i$. Each item in $\boldsymbol{y}^l_i$ consists of a semantic class $s$ (1-of-$C$ predefined classes), an amodal 3D bounding box parameterized by its center $c = (c^x, c^y, c^z)$, size $d = (d^x, d^y, d^z)$ and an orientation $\theta$ along the upright-axis.

\subsection{Teacher-Student Framework}
In semi-supervised learning, teacher-student framework~\cite{tarvainen2017mean} is widely used to propagate information from labeled data to unlabeled data. Here we introduce how the teacher-student framework works in the pseudo-label method. During the training stage, each batch contains labeled frames $\{\boldsymbol{x}^l_i\}^{B_l}_{i=1}$ and unlabeled frames $\{\boldsymbol{x}^u_i\}^{B_u}_{i=1}$, where $B_l$ and $B_u$ are the numbers of labeled frames and unlabeled frames in a batch, respectively.
The labeled frames $\{\boldsymbol{x}^l_i\}^{B_l}_{i=1}$ are used to supervise the student network as fully-supervised methods. 
For unlabeled frames $\{\boldsymbol{x}^u_i\}^{B_u}_{i=1}$, they are first feed to the teacher network. The predictions of the teacher network will be used as pseudo labels $\{\boldsymbol{\tilde{y}}^u_i\}_{i=1}^{N_u}$ of unlabeled frames. To ensure quality, the pseudo labels with low classification scores will be removed. As the training goes on, the teacher network generates pseudo labels for the student, which facilitates the learning of the student. The student gradually updates the teacher via exponential moving average (EMA), which makes a better teacher. Therefore, the teacher-student framework is trained in a mutually-beneficial manner.

\subsection{Gt Database and Pseudo Database}
\label{sec:gt-database}
Gt database and pseudo database are the bases of our semi-sampling. They collect ground truth samples and pseudo samples cropped from labeled frames and unlabeled frames. When using semi-sampling, we will randomly select object samples from the two databases.

\vspace{-2 mm}
\paragraph{Gt Database.} 
Gt-sampling is a data augmentation method introduced by \cite{yan2018second}, which is widely used in outdoor 3D object detection. It uses ground truth bounding boxes to crops ground truth samples from labeled frames, and the cropped object samples will be collected to generate a gt database $\mathcal{G} = \{s_i^l\}_{i=1}^{N_{gt}}$. $N_{gt}$ is the number of ground truth boxes in all labeled frames and $s_i^l$ denotes the $i^{th}$ ground truth sample. For $s_i^l$, it consists of the corresponding point clouds ${p}_i^l$ in labeled frames and its ground truth bounding box ${b}_i^l$. In the training stage, gt-sampling will randomly select ground truth samples from 
 $\mathcal{G}$ and pastes them to the current frame. To prevent pasted samples from overlapping with existing objects in the current frame, collision detection is needed.
 
For the KITTI dataset, we directly follow existing outdoor 3D object-detecting methods to build a gt database. For indoor scenes, as far as we know, we are the first to use gt-sampling. There are some differences between outdoor and indoor scenes. In indoor scenes, 3D object bounding boxes may overlap with each other, such as a chair under a table. Using ground truth bounding boxes to crop the gt samples will involve points from other objects, resulting in a noisy gt database. 
For the ScanNet\cite{dai2017scannet} dataset, as it provides an instance mask for each ground truth object, we can use instance masks instead of gt boxes to crop gt samples. In this way, only points belonging to the instance are cropped. 
For the SUN-RGBD\cite{song2015sun} dataset, there is no instance mask for each object, so we have to use ground truth bounding boxes to crop gt samples. As a result, the gt database of SUN-RGBD is not clean, which will produce unreal scenes when performing gt-sampling. In the experiments on SUN-RGBD, we find that gt-sampling can improve model performance at the early training epochs. However, we observe accuracy degradation at the end of training. We address this problem by using the fade strategy \cite{2021PointAugmenting}, which disables gt-sampling when the model is near convergent. Please refer to Table \ref{tab:indoor_gtsample} for more details. 

\vspace{-3 mm}
\paragraph{Pseudo Database.}
To build a pseudo database, we firstly pre-train a backbone detector on labeled frames $\{\boldsymbol{x}^l_i, \boldsymbol{y}^l_i\}_{i=1}^{N_l}$. Then, the pre-trained detector is used to forward unlabeled frames$\{\boldsymbol{x}^u_i\}_{i=1}^{N_u}$. The predictions are used as pseudo labels $\{\boldsymbol{\tilde{y}}^u_i\}_{i=1}^{N_u}$. Different from the gt database, we use \textit{pseudo labels} to crop pseudo samples from \textit{unlabeled frames}. With these pseudo samples, we can generate a pseudo database $\mathcal{P} = \{s_i^u\}_{i=1}^{N_{pse}}$, where $N_{pse}$ is the number of all pseudo samples and $s_i^u$ denotes the $i^{th}$ pseudo sample. For $s_i^u$, it consists of corresponding point clouds ${p}_i^u$ in unlabeled frames, predicted bounding box ${b}_i^{pred}$, and predicted classification score ${c}_i^{pred}$. Since there are many low-quality pseudo labels, when using the pseudo database, we will filter out high-quality pseudo samples by their scores.

\vspace{2 mm}
\subsection{Semi-Sampling}\label{sec:semi-sampling}
Current data augmentation methods are mainly designed for labeled data, but data augmentation for unlabeled data is less investigated. Moreover, unlabeled data can provide a large number of unseen objects which can be used for object sampling. Therefore, we propose semi-sampling to diversify labeled and unlabeled data and make full use of object samples from unlabeled data. As illustrated in Figure \ref{fig:overview}, our semi-sampling is composed of four object sampling strategies: gt-sampling on labeled frames, gt-sampling on unlabeled frames, pseudo-sampling on labeled frames, and pseudo-sampling on unlabeled frames. Here pseudo-sampling is sampling pseudo samples instead of gt samples compared to gt-sampling. Our semi-sampling method is a superset of the vanilla gt-sampling\cite{yan2018second} that only pastes gt samples to labeled frames.

\vspace{-2 mm}
\paragraph{Semi-Sampling on Labeled Frames.}
For gt-sampling on labeled frames, we follow the vanilla gt-sampling \cite{yan2018second}. For pseudo-sampling on labeled frames, we randomly select some pseudo samples along with their pseudo labels from the pseudo database. The pseudo labels of selected pseudo samples serve two purposes. Firstly, they are used to perform collision detection between pseudo samples and ground truth objects on labeled frames. Secondly, we use them as labels of pseudo samples in labeled frames to supervise the student network at the training time, which requires the pseudo-label to be of high quality. Therefore, we set a pseudo label score threshold $\tau_{pseudo\_sample}$. When performing pseudo-sampling, we only use pseudo samples whose pseudo label scores $c > \tau_{pseudo\_sample}$.
\vspace{-2 mm}
\paragraph{Semi-Sampling on Unlabeled Frames.}
When employing semi-sampling on unlabeled frames, there are no ground truth bounding boxes on unlabeled frames for collision detection with pasted samples.
Considering that pseudo labels can provide approximate localization of foreground objects in unlabeled frames,  we directly use them for collision detection. However, the massive pseudo labels may take up quite a lot of space. So we set a threshold $\tau_{unlabeled\_frame}$ to filter the pseudo labels in unlabeled frames with pseudo label scores $c > \tau_{unlabeled\_frame}$. It is worth noting that the ground truth labels and pseudo labels of the pasted samples on the unlabeled frames will not be used to supervise the student network. All supervising information of unlabeled frames is produced by the teacher network online.

\vspace{-2 mm}
\paragraph{Category Shuffling.}
When performing semi-sampling, we select several samples for \textit{each category} from the gt database or pseudo database. However, indoor scenes are crowded and there are many categories, such as 18 for ScanNet and 10 for SUN-RGBD. If we choose samples according to a fixed category queue like outdoor methods, there may be no space for the samples of the category at the end of the queue, which will cause inter-class sample imbalance. To solve the issue, we simply shuffle the category queue every time we use semi-sampling.

\section{Experiments}
\subsection{Datasets and Evaluation Metrics}
\paragraph{ScanNet and SUN-RGBD.} For indoor scenes, we evaluate our method on ScanNet\cite{dai2017scannet} and SUN-RGBD\cite{song2015sun}. ScanNet is an indoor dataset consisting of 1,513 reconstructed meshes, among which 1,201 are training samples and the rest are validation samples. SUN-RGBD contains 10,335 RGB-D images of indoor scenes, which are split into 5,285 training samples and 5,050 validation samples. We follow a standard evaluation protocol~\cite{qi2019deep} by using mean Average Precision (mAP) under iou thresholds of 0.25 and 0.50.

\vspace{-2 mm}
\paragraph{KITTI.} For outdoor scenes,  we use KITTI\cite{geiger2013vision} for evaluation. KITTI is a popular dataset for autonomous driving, which consists of fine annotations for 3D object detection. There are 7481 outdoor scenes for training and 7518 for testing. The training samples are generally divided into a training split of 3712 samples and a validation split of 3769 samples. We report the mAP with 40 recall positions, with a rotated IoU threshold of 0.7, 0.5, and 0.5 for the three classes, car, pedestrian, and cyclist, respectively.

\vspace{-2 mm}
\paragraph{Omni-KITTI.}
Omni-supervised learning\cite{radosavovic2018data} is a special regime of semi-supervised learning, which uses all labeled data as well as large-scale unlabeled data. It is lower-bounded by the accuracy of training on
all labeled data and it aims to boost the fully-supervised methods. To validate the effectiveness of our method under the omni-supervised setting, we construct the Omni-KITTI dataset with KITTI raw data. Specifically, KITTI raw data consists of many sequences containing all frames that emerge on the KITTI training and validation set. Therefore, we can use the KITTI training set as all labeled data (including 3712 samples) and take the remaining data that excludes KITTI validation sequences as unlabeled data (including 33507 samples). The final performance is evaluated on the KITTI validation set with the same evaluation protocol as the KITTI dataset.

\newcommand{\tabincell}[2]{\begin{tabular}{@{}#1@{}}#2\end{tabular}}  
\begin{table*}[t]
    \begin{center}
    \scalebox{0.9}[0.9]{
        \begin{tabular}{c|c|cc|cc|cc|cc}
            \hline
            & & \multicolumn{2}{c|}{5\%} & \multicolumn{2}{c|}{10\%} & \multicolumn{2}{c|}{20\%} & \multicolumn{2}{c}{100\%} \\ 
            \cline{3-10}

            \multirow{-2}{*}{Dataset} & \multirow{-2}{*}{Model}   & \tabincell{c}{mAP\\ @0.25} & \tabincell{c}{mAP\\@0.5} & \tabincell{c}{mAP\\@0.25} & \tabincell{c}{mAP\\@0.5} & \tabincell{c}{mAP\\@0.25} & \tabincell{c}{mAP\\@0.5} & \tabincell{c}{mAP\\@0.25} & \tabincell{c}{mAP\\@0.5} \\ 
            \hline
            \hline
            
            \multirow{5}{*}{ScanNet} & VoteNet\cite{qi2019deep}  &27.9	&10.8	&36.9	&18.2	&46.9	&27.5	&57.8	&36.0 \\ \cline{2-10} 
            & SESS\cite{zhao2020sess}  & $\backslash$ & $\backslash$ &39.7	&18.6	&47.9	&26.9	&62.1	&38.8  \\ \cline{2-10} 
            & 3DIoUMatch\cite{wang20213dioumatch} &40.0	&22.5	&47.2	&28.3	&52.8	&35.2	&62.9	&42.1 \\ \cline{2-10} 
            & Semi-Sampling &\textbf{41.1}	&\textbf{27.7}	&\textbf{50.3}	&\textbf{34.7}	&\textbf{54.4}	&\textbf{38.9}	&\textbf{64.6}	&\textbf{47.7}  \\ \cline{2-10} 
            & \textit{improvements} &\textit{+1.1}	&\textit{+5.2}	&\textit{+3.1}	&\textit{+6.4}	&\textit{+1.6}	&\textit{+3.7}	&\textit{+1.7}	&\textit{+5.6} \\ \cline{2-10} 
            \hline
            \multirow{5}{*}{SUN-RGBD} & VoteNet\cite{qi2019deep} &29.9	&10.5	&38.9	&17.2	&45.7	&22.5	&58.0	&33.4\\ \cline{2-10}
            & SESS\cite{zhao2020sess}  & $\backslash$ & $\backslash$ &42.9	&14.4	&47.9	&20.6	&61.1	&37.3 \\ \cline{2-10}
            & 3DIoUMatch\cite{wang20213dioumatch} &39.0	&21.1	&45.5	&28.8	&49.7	&30.9	&61.5	&41.3 \\  \cline{2-10} 
            & Semi-Sampling & \textbf{40.1}	&\textbf{24.3}	&\textbf{50.0}	&\textbf{31.5}	&\textbf{54.4}	&\textbf{34.5}	&\textbf{63.2}	&\textbf{45.7} \\ \cline{2-10}
            & \textit{improvements} &\textit{+1.1} &\textit{+3.2} &\textit{+4.5} &\textit{+2.7} &\textit{+4.7} &\textit{+3.6} &\textit{+1.7} &\textit{+4.4} \\ \cline{2-10} 
            \hline
        \end{tabular}
    }
    \end{center}
    \vspace{-2mm}
    \caption{Comparison between Semi-Sampling and state-of-the-arts on ScanNet and SUN-RGBD val set under different ratios of labeled data. Following prior works, we also report the results with 100\% labeled data, where the unlabeled data is a copy of the full dataset.}
    \label{tab:indoor_main}
\end{table*}

\begin{table*}[t]
    \begin{center}
    \scalebox{0.9}[0.9]{
        \begin{tabular}{c|ccc|ccc|ccc}
        \hline
        & \multicolumn{3}{c|}{1\%} & \multicolumn{3}{c|}{2\%} & \multicolumn{3}{c}{5\%}\\ 
          \cline{2-10} \multirow{-2}{*}{Model} & Car & Pedestrian & Cyclist & Car & Pedestrian & Cyclist & Car &Pedestrian & Cyclist  \\ 
        \hline
        \hline
        PV-RCNN\cite{shi2020pv} & 75.0 & 16.1 & 39.2 & 78.5 & 38.2 & 75.0 & 80.0 & 52.8 & 56.9 \\ \cline{1-10} 
        3DIoUMatch\cite{wang20213dioumatch} & 75.1 & 17.4 & 44.1 & 77.9 & 33.4 & 52.4 & 82.0 & 52.6 & 64.2 \\ \cline{1-10}
        Semi-Sampling & \textbf{78.6} & \textbf{24.1} & \textbf{58.2} & \textbf{81.9} & \textbf{47.5} & \textbf{63.3} & \textbf{82.3} & \textbf{54.9} & \textbf{71.1} \\ \cline{1-10} 
        \textit{improvements} & \textit{+3.5} & \textit{+6.7} & \textit{+14.1} & \textit{+4.0} & \textit{+14.1} & \textit{+10.9} & \textit{+0.3} & \textit{+2.3} & \textit{+6.9} \\ \cline{1-10} 
        \end{tabular}
    }
    \end{center}
    \vspace{-2mm}
    \caption{Comparison between PV-RCNN, 3DIoUMatch and Semi-Sampling on KITTI val set under different ratios of labeled data. The results are evaluated
    with the AP calculated by 40 recall positions for car, pedestrian and cyclist classes. All the models in the table are trained with the same data split under each ratio.}
    \label{tab:kitti_main}
    \vspace{-2mm}
\end{table*}

\subsection{Implementation Details}
\paragraph{ScanNet and SUN-RGBD.}
We re-implement the indoor 3DIoUMatch~\cite{wang20213dioumatch} on the OpenPCDet\cite{openpcdet2020}, achieving similar results as the official repo. Then we use the 3DIoUMatch as our baseline. In 3DIoUMatch, the backbone detector is an IoU-aware VoteNet which is a VoteNet equipped with a 3D IoU estimation module. Our backbone detector is also the same as 3DIoUMatch. All the experiments are trained on 4 GPUs.
\textit{At pre-training stage}, we train the IoU-aware VoteNet with a batch size of 8 and use the same data augmentation as 3DIoUMatch\cite{wang20213dioumatch}. The IoU-aware VoteNet is optimized by an adam optimizer with an initial learning rate of 0.008 for 900 epochs. The learning rate is decayed by a factor of 0.1 at the $400^{th}$, $600^{th}$, $800^{th}$ epoch. The pre-trained model is not only used for the initialization of the training stage but also used to generate pseudo labels for unlabeled data and build a pseudo database. \textit{At the training stage}, we train our method with a batch size of $\{4+8\}$, where 4 is the number of labeled frames and 8 is the number of unlabeled frames. For the ScanNet dataset, our model is optimized by an adam optimizer with an initial learning rate of 0.004 for 1000 epochs. We decay the learning rate by 0.3, 0.3, 0.1, 0.1 at the $400^{th}$, $600^{th}$, $800^{th}$, $900^{th}$ epoch, following the training strategy of 3DIoUMatch. For the SUN-RGBD dataset, the training is time-consuming. To speed up the convergence, we train our model for 300 epochs with an adam one-cycle optimizer and a maximum learning rate of 0.008. \textit{At the inference stage}, we use the student IoU-aware VoteNet to forward input data.

\vspace{-3mm}
\paragraph{KITTI.}
There is an official implementation of outdoor 3DIoUMatch on OpenPCDet. We can directly use it as our baseline.
We employ PV-RCNN as our backbone detector following 3DIoUMatch. 
\textit{At pre-training stage}, the backbone detector is optimized by an adam one-cycle optimizer with the batch size 8, learning rate 0.01 for 80 epochs. For all ratios of labeled data, we repeat the labeled data 10 times. 
\textit{At the training stage}, we repeat the labeled data 5 times and train the model for 60 epochs with the same optimizer and learning rate as the pre-training stage. For the experiments under 1\% and 2\% labeled data, we set $\tau_{unlabeled\_frame}$ = 0.5 to filter abundant pseudo labels on unlabeled frames for collision detection. The $\tau_{pseudo\_sample}$, which is used for selecting pseudo samples for pseudo-sampling, is set to 0.8, 0.7, 0.7 for car, pedestrian and cyclist, respectively. For the experiments under 5\% labeled data, we use higher $\tau_{unlabeled\_frame}$ and $\tau_{pseudo\_sample}$. Because 
 with more labeled data, the pre-trained model will produce more high-quality pseudo labels, which usually have higher scores. In practice, we set $\tau_{unlabeled\_frame}$ to 0.8 and $\tau_{pseudo\_sample}$ to 0.9, 0.8, 0.8 for car, pedestrian and cyclist.
\textit{At the inference stage}, we use the student PV-RCNN network to forward input data.

\subsection{Comparison with State-of-the-Arts}
\paragraph{ScanNet and SUN-RGBD.}
We compare our methods with previous state-of-the-art works SESS\cite{zhao2020sess} and 3DIoUMatch\cite{wang20213dioumatch} on ScanNet and SUN-RGBD dataset under 5\%, 10\%, 20\% and 100\% labeled data. Our method outperforms the current state-of-the-art, 3DIoUMatch\cite{wang20213dioumatch}, under all ratios of labeled data and all listed datasets. For example, with the same backbone detector and the same split of 10\% labeled data, our approach achieves 50.3 mAP@0.25 and 34.7 mAP@0.5 on ScanNet, outperforming previous best results by 3.1 and 6.4 points. For the SUN-RGBD dataset, we also attain decent results. With the same backbone detector and split of 5\% labeled data, we improve 3DIoUMatch by 1.1 and 3.2 points in terms of mAP@0.25 and mAP@0.5.
When utilizing 100\% labeled data, there are still gains on two datasets, suggesting that semi-supervised learning is powerful with only labeled data. 
It should be noted that because the gt and pseudo databases of the SUN-RGBD dataset are noisy, we do not use semi-sampling when training 10\%, 20\% and 100\% SUN-RGBD. Instead, we perform gt-sampling with the fade strategy\cite{2021PointAugmenting} on the pre-trained model. The good pre-trained models can also benefit the semi-supervised learning on 10\%, 20\% and 100\% SUN-RGBD, achieving promising results.

\vspace{-1mm}
\paragraph{KITTI.}
We also provide a comparison with 3DIoUMatch on the outdoor dataset KITTI. As shown in Table \ref{tab:kitti_main}, our semi-sampling method yields significant improvement. In particular, we outperform 3DIoUMatch by 3.5 mAP, 6.7 mAP and 14.1 mAP on car, pedestrian and cyclist classes under 1\% labeled data. With only 5\% labeled data, the performance of our method is close to the methods trained with 100\% labeled data, as can be seen in Table \ref{tab:kitti_ablation3}.

\subsection{Ablation Study}
In this section, we conduct extensive ablation experiments to analyze the effect of our semi-sampling on three widely used 3D object detection datasets.

\vspace{-1mm}
\paragraph{Effect of Semi-Sampling on KITTI.}
We first ablate the effectiveness of each object sampling strategy of semi-sampling on the KITTI dataset.
Our semi-sampling consists of gt-sampling on labeled frames, gt-sampling on unlabeled frames, pseudo-sampling on labeled frames, and pseudo-sampling on unlabeled frames. In outdoor scenes, the effectiveness of gt-sampling on labeled frames has been verified by a lot of prior works, so we use it by default.
All experiments in Table \ref{tab:kitti_ablation1} use the same pre-trained weight.
In Table \ref{tab:kitti_ablation1}, experiment (a) is our baseline, namely 3DIoUMatch. Experiment (b) suggests that the pseudo-sampling on labeled frames can improve our baseline by a large margin. Further, 

\begin{table}[t]
    \begin{center}
    \scalebox{0.75}[0.75]{
        \begin{tabular}{c|cc|cc|ccc}
        \hline
         \multirow{2}{*}{Exp.} & \multicolumn{2}{c|}{Labeled Frame} & \multicolumn{2}{c|}{Unlabeled Frame} & \multicolumn{3}{c}{KITTI 1\%}\\
         \cline{2-8}
        & gt. & pseudo. & gt. & pseudo. & Car & Pedestrian & Cyclist \\ 
        \hline
        \hline
        (a) & \checkmark & & & & 75.1 & 17.4 & 44.1 \\ \hline
        (b) & \checkmark & \checkmark & & & 77.6 & 20.3 & 56.0 \\ \hline
        (c) & \checkmark & \checkmark & \checkmark & & 77.3 & 17.1 & 55.6 \\ \hline
        (d) & \checkmark & \checkmark & & \checkmark & 78.5 & 21.0 & 57.8 \\ \hline
        (e) & \checkmark & \checkmark & \checkmark & \checkmark & \textbf{78.6} & \textbf{24.1} & \textbf{58.2} \\ \hline
        \end{tabular}
    }
    \end{center}
    \vspace{-4mm}
    \caption{Ablation of semi-sampling on KITTI 1\% labeled data. The results are for moderate difficulty level evaluated by the 3D mAP with 40 recall positions. ``gt.'' and ``pseudo.'' represent gt-sampling and pseudo-sampling.}
    \label{tab:kitti_ablation1}
\end{table}

\begin{table}[h]
    \begin{center}
    \scalebox{0.75}[0.75]{
        \begin{tabular}{c|cc|cc|cc}
        \hline
        \multirow{2}{*}{Exp.} & \multicolumn{2}{c|}{Labeled Frame} & \multicolumn{2}{c|}{Unlabeled Frame} & \multicolumn{2}{c}{ScanNet 10\%}\\
        \cline{2-7}
        & gt. & pseudo. & gt. & pseudo. & mAP@0.25 & mAP@0.50 \\ 
        \hline
        \hline
        (a) & & & & & 47.3 &29.6 \\ \hline
        (b) & \checkmark & & & & 48.3	&32.7 \\ \hline
        (c) & \checkmark & \checkmark & & & 47.7 &31.7 \\ \hline
        (d) & \checkmark &  & \checkmark & & \textbf{50.3}	&\textbf{34.7} \\ \hline
        (e) & \checkmark &  & & \checkmark & 48.2	&33.1\\ \hline
        (f) & & \checkmark & & 	&47.6 &30.4 \\ \hline
        (g) & &  & \checkmark & &46.7 &29.1 \\ \hline
        (h) & &  & & \checkmark &46.4 &28.5\\ \hline
        \end{tabular}
    }
    \end{center}
    \vspace{-4mm}
    \caption{Ablation study of semi-sampling on ScanNet 10\% labeled data. We use the 10\% data split provided by 3DIoUMatch as labeled data for a fair comparison.}
    \label{tab:indoor_ablation1}
\end{table}

\begin{table}[h]
    \begin{center}
    \scalebox{0.75}[0.75]{
        \begin{tabular}{c|cc|cc|cc}
        \hline
        \multirow{2}{*}{Exp.} & \multicolumn{2}{c|}{Labeled Frame} & \multicolumn{2}{c|}{Unlabeled Frame} & \multicolumn{2}{c}{SUN-RGBD 5\%}\\
        \cline{2-7}
        & gt. & pseudo. & gt. & pseudo. & mAP@0.25 & mAP@0.50 \\ 
        \hline
        \hline
        (a) & & & & & 39.4	&23.0 \\ \hline
        (b) & \checkmark & & & & 40.0	&23.5 \\ \hline
        (c) & \checkmark & \checkmark & & & 40.0	&22.8 \\ \hline
        (d) & \checkmark &  & \checkmark & & \textbf{40.1}	&\textbf{24.3} \\ \hline
        (e) & \checkmark &  & & \checkmark & 39.1	&24.2\\ \hline
        (f) & & \checkmark & & 	&39.4 &23.9 \\ \hline
        (g) & &  & \checkmark & &39.2 &22.9 \\ \hline
        (h) & &  & & \checkmark &40.0 &22.9\\ \hline
        \end{tabular}
    }
    \end{center}
    \vspace{-4mm}
    \caption{Ablation study of semi-sampling on SUN-RGBD 5\% labeled data. We use the 5\% data split provided by 3DIoUMatch as labeled data for a fair comparison.}
    \label{tab:indoor_ablation2}
    \vspace{-4mm}
\end{table}

\noindent
experiment (d) obtains 0.9, 0.7, 1.8 improvements on car, pedestrian and cyclist compared to experiment (b), demonstrating the effectiveness of pseudo sampling on unlabeled frames. Interestingly, gt sampling on unlabeled frames leads to a decrease to experiment (b) as seen in experiment (c), while leading to an increase to experiment (d) as seen in experiment (e). Overall, when equipped with all object sampling strategies in our semi-sampling, we achieve the best performance, which indicates that it is helpful to augment both labeled and unlabeled data with the rich object samples from both of them. 

\paragraph{Effect of Semi-Sampling on ScanNet and SUN-RGBD.}
In order to be consistent with 3DIoUMatch, we ablate our method on ScanNet with \textit{10\% labeled data} and on SUN-RGDB with \textit{5\% labeled data}. We use pre-trained weights provided by 3DIoUMatch on ScanNet 10\% and SUN-RGBD 5\%  to initialize our model in Table \ref{tab:indoor_ablation1} and Table \ref{tab:indoor_ablation2} for a fair comparison. And we use 3DIoUMatch as our baseline, corresponding to experiment (a) in both tables.
Experiment (b) in Table \ref{tab:indoor_ablation1} and Table \ref{tab:indoor_ablation2} demonstrate the effectiveness of gt-sampling on labeled frames.
Experiment (f)-(h) in Table \ref{tab:indoor_ablation1} suggest that pseudo-sampling on labeled frames is also helpful while gt-sampling or pseudo-sampling on unlabeled frames can not lead to improvement on ScanNet. By contrast, Experiment (f)-(h) in Table \ref{tab:indoor_ablation2} shows that pseudo-sampling on labeled and unlabeled frames benefits our baseline while gt-sampling on unlabeled frames can not bring improvement on SUN-RGBD. Overall, the best performance on both datasets is achieved when using gt-sampling on labeled and unlabeled frames.

\vspace{2mm}
\begin{figure}[h]
\begin{center}
  \includegraphics[width=0.8\linewidth]{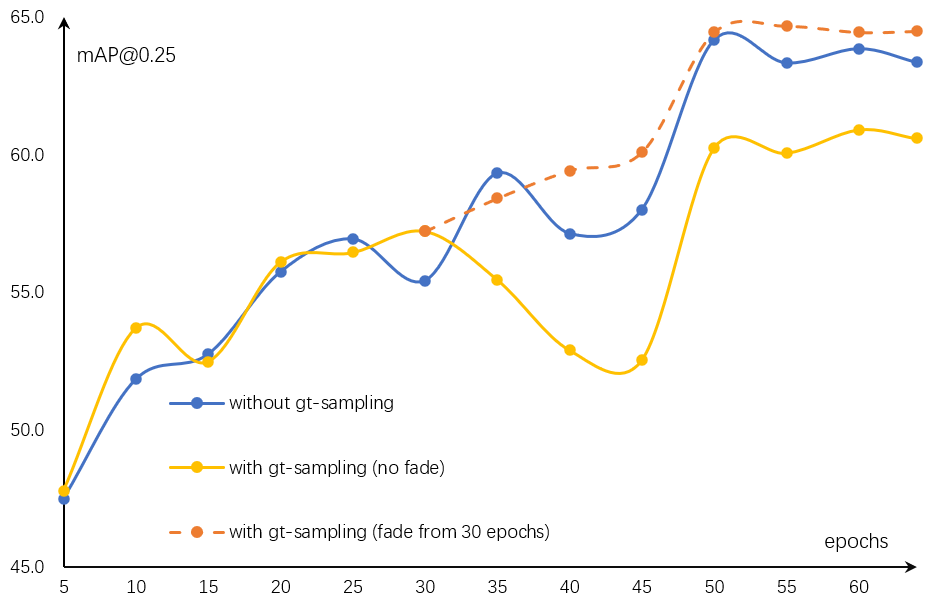}
\end{center}
\vspace{-3mm}
  \caption{Performance of RBGNet on SUN-RGBD.}
\label{fig:gt-sampling_sunrgbd}
\vspace{-4mm}
\end{figure}

\begin{table}[h]
    \begin{center}
    \scalebox{0.7}[0.7]{
        \begin{tabular}{c|c|c|cc|cc}
            \hline
            & & &\multicolumn{2}{c|}{ScanNet 100\%} & \multicolumn{2}{c}{SUN-RGBD 100\%} \\ 
            \cline{4-7}

            \multirow{-2}{*}{Exp.} & \multirow{-2}{*}{Gt-sampling} & \multirow{-2}{*}{Fade Strategy}     & \tabincell{c}{mAP\\ @0.25} & \tabincell{c}{mAP\\@0.5} & \tabincell{c}{mAP\\@0.25} & \tabincell{c}{mAP\\@0.5} \\
            \hline
            \hline 
            (a) &            & &70.2	&54.2	&64.1	&47.2 \\\hline
            (b) & \checkmark & &\textbf{71.1}	&\textbf{56.4}	&61.0	&46.4   \\ \hline 
            (c) & \checkmark & from $20^{th}$ epoch &70.7	&56.0	&64.6	&48.1   \\ \hline
            (d) & \checkmark & from $30^{th}$ epoch &70.2	&55.5	&\textbf{65.1}	&\textbf{48.5}   \\ \hline
            (e) & \checkmark & from $40^{th}$ epoch &71.1	&55.7	&64.9	&47.5   \\ \hline
            \hline
        \end{tabular}
    }
    \end{center}
    \vspace{-3mm}
    \caption{Ablation of gt-sampling on RBGNet. The total epoch of RBGNet is 64. ``from $20^{th}$ epoch'' means that we disable gt-sampling after the $20^{th}$ epochs, and so on.}
    \label{tab:indoor_gtsample}
    \vspace{-4mm}
\end{table}

\vspace{-2mm}
\paragraph{Effect of Gt-Sampling on Fully-Supervised Method.}
\label{sec:fade-strategy}
As discussed in Section \ref{sec:gt-database}, gt-sampling is a very useful data augmentation method in \textit{outdoor scenes} while there is no investigation on it in \textit{indoor scenes}. Here we conduct an experiment to study the effectiveness of gt-sampling in indoor scenes.
To provide more value to the community, we use a fully-supervised state-of-the-art method RBGNet\cite{wang2022rbgnet} as our baseline.
As shown in experiment (b) of Table \ref{tab:indoor_gtsample}, gt-sampling boosts the strong baseline by 0.9 mAP@0.25 and 2.2 mAP@0.5 on ScanNet, achieving a new state-of-the-art on ScanNet. However, there is a decrease on SUN-RGBD. As depicted in Figure \ref{fig:gt-sampling_sunrgbd}, in the early stage, gt-sampling is actually beneficial to the model performance. However, the performance degrades at the end compared to the baseline. We attribute the phenomenon to the noisy gt database of SUN-RGBD as mentioned in Section \ref{sec:gt-database}. In the early epochs, the model is not well-trained, and gt-sampling can augment the training data largely. In the late epochs, cleaner data are needed to refine the well-trained model, so the noisy gt samples become harmful for training. To make use of gt-sampling on SUN-RGBD, we use the fade strategy \cite{2021PointAugmenting}, which disables gt-sampling when the model is near convergent. We use three different epochs to study the best time to fade gt-sampling. As shown in experiments (c)-(e), when fading from the $30^{th}$ epoch, we get the best result and achieve a new state-of-the-art on SUN-RGBD.
\vspace{2mm}
\begin{table}[h]
    \begin{center}
    \scalebox{0.85}[0.85]{
        \begin{tabular}{c|ccc}
        \hline
         & \multicolumn{3}{c}{3D mAP} \\ \cline{2-4}
        \multirow{-2}{*}{\tabincell{c}{$\tau_{unlabeled\_frame}$}} & Car & Pedestrian & Cyclist \\ \hline \hline
        0.3 & 78.4 & 19.2 & 57.4 \\ \hline
        0.5 & \textbf{78.6} & \textbf{24.1} & \textbf{58.2} \\ \hline
        0.7 & 78.0 & 20.0 & 57.0 \\ \hline
        \end{tabular}
    }
    \end{center}
    \vspace{-4mm}
    \caption{Further study on the influence of the threshold $\tau_{unlabeled\_frame}$ on KITTI 1\% labeled data. The results are for moderate difficulty evaluated by the mAP with 40 recall positions.}
    \label{tab:kitti_ablation2}
    \vspace{-6mm}
\end{table}

\paragraph{Pseudo Label Threshold on Unlabeled Frames.}
When performing semi-sampling on unlabeled frames, we use pseudo labels of unlabeled frames to do collision detection with the pasted samples. To remove low-quality pseudo labels in unlabeled frames, we set a score threshold $\tau_{unlabeled\_frame}$ to filter them.
A low threshold will leave too many abundant pseudo labels, making there no space on unlabeled frames to place object samples.
A high threshold will filter pseudo labels of some foreground objects on unlabeled frames. Without pseudo labels for collision detection, the pasted samples may overlap with them. In Table \ref{tab:kitti_ablation2}, we study three different thresholds, and the experiment shows that 0.5 is a good choice for our method.

\begin{figure*}[t]
	\vspace{-4pt}
	\begin{center}
		\setlength{\fboxrule}{0pt}
		\fbox{\includegraphics[width=0.97\textwidth]{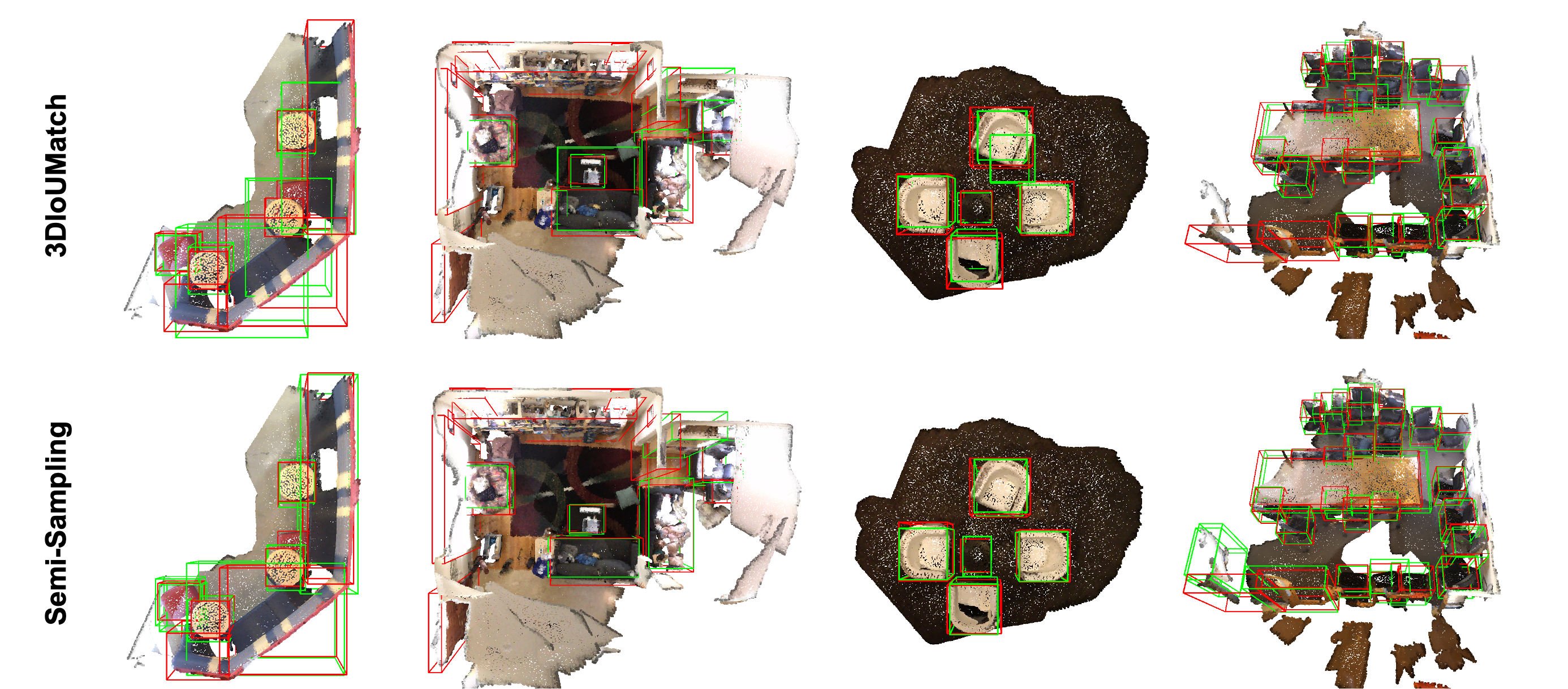}}
	\end{center}
	\captionsetup{font={small}}
	\vspace{-8pt}
	\caption{Qualitative results on ScanNet with 10\% labeled data. We show ground truth boxes and predictions in red and green. }
	\label{fig:vis_scannet}
	\vspace{-2pt}
\end{figure*}

\begin{figure*}[t]
	\vspace{-4pt}
	\begin{center}
		\setlength{\fboxrule}{0pt}
		\fbox{\includegraphics[width=0.97\textwidth]{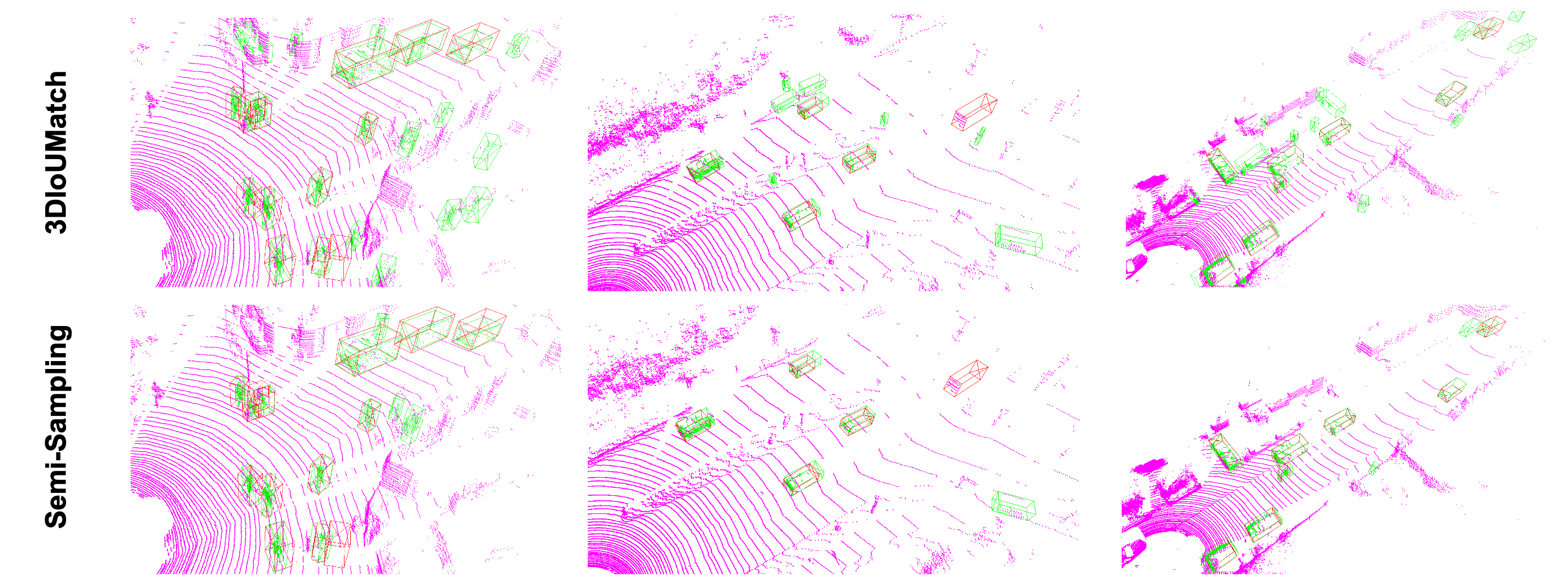}}
	\end{center}
	\captionsetup{font={small}}
	\vspace{-8pt}
	\caption{Qualitative results on KITTI with 1\% labeled data. We show ground truth boxes and predictions in red and green. }
	\label{fig:vis_kitti}
	\vspace{-4pt}
\end{figure*}

\paragraph{Performance on Omni-KITTI.}
\begin{table}[h]
    \begin{center}
    \vspace{-3mm}
    \scalebox{0.82}[0.82]{
        \begin{tabular}{c|ccc}
        \hline
        & & Omni-KITTI \\
        \cline{2-4}
        \multirow{-2}{*}{\tabincell{c}{Method}}  & Car & Pedestrian & Cyclist \\ \hline \hline
        PV-RCNN\cite{shi2020pv} & 84.4	&54.5	&70.4 \\ \hline
        PV-RCNN + 3DIoUMatch & 84.3 & 55.1 & 71.9 \\ \hline
        PV-RCNN + Semi-Sampling & \textbf{85.2} &\textbf{64.2} &\textbf{74.3} \\ \hline \hline
        CT3D\cite{sheng2021improving} & 85.0	&55.6	&71.9 \\ \hline
        CT3D + Semi-Sampling & \textbf{85.8} & \textbf{64.8} & \textbf{72.8} \\ \hline
        \end{tabular}
    }
    \end{center}
    \vspace{-4mm}
    \caption{Further study on KITTI 100\% labeled data with Omni-KITTI as unlabeled data.}
    \label{tab:kitti_ablation3}
\end{table}

To further verify the effectiveness of our method under the omni-supervised setting, we evaluate our method on the Omni-KITTI dataset.
We use the pre-train weight provided by PV-RCNN to initialize 3DIoUMatch and our method. Table \ref{tab:kitti_ablation3} summarize the results. Our method pushes the fully-supervised PV-RCNN to a new level. Specifically, we outperform PV-RCNN by 0.8 mAP, 9.7 mAP and 3.9 mAP on car, pedestrian and cyclist classes. For 3DIoUMatch, its improvement on PV-RCNN is relatively small. In addition, we use another backbone detector CT3D\cite{sheng2021improving} to validate our method.
We use the pre-train weight provided by CT3D as initialization. As shown in Table \ref{tab:kitti_ablation3}, the improvement is also considerable, especially on the pedestrian class.

\subsection{Qualitative Results and Analysis}
Here we provide the qualitative comparison between our method and 3DIoUMatch on ScanNet and KITTI validation set.
For better visualization on ScanNet, we filter out the predictions whose classification scores $s > 0.25$.
As shown in Figure \ref{fig:vis_scannet} and Figure \ref{fig:vis_kitti}, our method can suppress false positives greatly.
Moreover, our method gives more accurate predictions and recalls more challenging objects, such as the chairs under the table, which confirms the effectiveness of our semi-sampling again. More qualitative results are provided in the supplementary material.

\vspace{-2mm}
\section{Conclusion}
In this paper, we propose semi-sampling, a simple but effective data augmentation method for semi-supervised 3D object detection. 
In general, we perform object sampling on both labeled and unlabeled frames and utilize unlabeled data to augment labeled data. In addition, we contribute a new benchmark dataset setting Omni-KITTI for the study of omni-supervised learning. Experiments on the ScanNet, SUN-RGBD, KITTI and Omni-KITTI datasets demonstrate the effectiveness of our method.

{\small
\bibliographystyle{ieee_fullname}
\bibliography{egbib}
}

\end{document}